# Developing a Hybrid Convolutional Neural Network for Automatic Aphid Counting in Sugar Beet Fields

Xumin Gao [b,e], Wenxin Xue [a,*], Callum Lennox [b,e], Mark Stevens [c], Junfeng Gao [d,e,*]

[a] Tobacco Research Institute, Chinese Academy of Agricultural Sciences, Qingdao, 266101, China

[b] School of Computer Science, University of Lincoln, Lincoln, LN6 7TS, UK

[c] British Beet Research Organisation, Norwich Research Park, Innovation Centre, Colney Lane, Norwich, NR4 7GJ, UK

[d] Lincoln Agri-Robotics Centre, Lincoln Institute for Agri-Food Technology, University of Lincoln, Lincoln, LN2 2LG, UK

[e] Lincoln Centre for Autonomous Systems (L-CAS), University of Lincoln, Lincoln, LN6 7TS, UK

**ABSTRACT**

Aphids can cause direct damage and indirect virus transmission to crops. Timely monitoring and control of their populations are thus critical. However, the manual counting of aphids, which is the most common practice, is labor-intensive and time-consuming. Additionally, two of the biggest challenges in aphid counting are that aphids are small objects and their density distributions are varied in different areas of the field. To address these challenges, we proposed a hybrid automatic aphid counting network architecture which integrates the detection network and the density map estimation network. When the distribution density of aphids is low, it utilizes an improved Yolov5 to count aphids. Conversely, when

* Corresponding author.

*E-mail address*: jugao@lincoln.ac.uk (J. Gao), xuewenxin@caas.cn (W. Xue)

the distribution density of aphids is high, it switches to CSRNet to count aphids. To the best of our knowledge, this is the first framework integrating the detection network and the density map estimation network for counting tasks. Through comparison experiments of counting aphids, it verified that our proposed approach outperforms all other methods in counting aphids. It achieved the lowest MAE and RMSE values for both the standard and high-density aphid datasets: 2.93 and 4.01 (standard), and 34.19 and 38.66 (high-density), respectively. Moreover, the AP of the improved Yolov5 is 5% higher than that of the original Yolov5. Especially for extremely small aphids and densely distributed aphids, the detection performance of the improved Yolov5 is significantly better than the original Yolov5. This work provides an effective early warning caused by aphids in sugar beet fields, offering protection for sugar beet growth and ensuring sugar beet yield. The datasets and project code are released at: https://github.com/JunfengGaolab/Counting-Aphids.

# 1. Introduction

As one of the agricultural pests, aphids can seriously affect the growth and yield of agricultural crops. They cause direct damage to crops through feeding behavior. Additionally, some aphids could transit virus by feeding, leading to indirect damage and yield loss. In 2020, the Fenland area of Cambridgeshire, UK, witnessed widespread infection in various fields due to the yellows virus, primarily transmitted by the peach potato aphid. The infection rates soared to 100%. This ultimately led to 38.1% of crops in the UK being infected with the virus yellows (Dewar and Qi, 2021). In Australia, the potential crop economic losses caused by aphid feeding and virus transmission are $241million and $482 million per year respectively (Valenzuela and Hoffmann, 2015). In Europe, sugarbeet virus yellows transmitted by aphids caused a 49% reduction in sugar yield of beets (Dedryver et al., 2010). These figures highlight the significant impact of aphids on crop production and the urgent need for effective pest management strategies.

To control the polulation of aphids, the most common practice is to spray pesticides on crops. However, this will lead to excessive use of pesticides due to the short breeding cycle of aphids. To rationalize and target the use of pesticides, representative sampling methods have been used as a preventive measure to limit the transmission source. Specifically, yellow water pan traps are placed in different areas of the fields to attract and trap aphids. Then the samples are sent to laboratories for manual counting under a microscope by entomologists. Finally, the counting results and early warnings are published to farmers, prompting them to target pesticides on the areas where aphids are likely to break out at a large scale. This method helps cut off the source of aphids and prevent the spread of virus yellows caused by aphids to other regions (Behrens, 2010). However, the manual counting of aphids is labor-intensive and time-consuming. And it requires expert knowledge to differentiate aphids from other insects.

There are some existing works to explore automatic aphid counting solutions. These works can be broadly categorized into three types: image processing, traditional machine learning, and deep learning. For the work based on image processing, it mainly uses some traditional image processing techniques for aphid identification. Suo et al. (2017) combined Grabcut foreground segmentation (Rother et al., 2004) and Otsu threshold segmentation (Otsu, 1979) to segment aphids from yellow pasteboards, and then count the aphids by extracting the contour and area screening. Shajahan et al. (2017) used watershed segmentation and edge detection to preliminarily extract all of pests from the images. They then designed a set of shape classification parameters to identify and count aphids of interest with greater than 82.4% accuracy on the consumer-grade digital camera. In terms of traditional machine learning, the main concept is to design hand-crafted features based on color, shape, texture and other features of targets, and feed these extracted features into the machine learning classifier for training and test. Liu et al. (2016) established a database including positive and negative samples of aphids. Then, they recognized aphids by extracting the histogram of oriented gradient (HOG) (Dalal and Triggs, 2005) feature and using a support vector machine (SVM) classifier (Platt, 1998), achieving 86.81% accuracy. Moreno-Lucio et al. (2021) filtered out the region of interest by extracting the color and morphological features of all of insects in the images. They then generated descriptors for classification by extracting the scale-invariant feature transform (SIFT) (Lowe, 1999) features of each class of insects, including aphids and other pests. Their experimental results showed that the average classification precision of the algorithm is 90%. For the work based on deep learning, it uses deep convolution layers to automatically extract the feature maps of aphids and classify them. There are two branches including the detection network and the density map estimation network. Li et al. (2019) proposed a two-stage CNN (Convolutional Neural Network) that operates in a coarse-to-fine manner. The coarse network initially determines the location of aphids or aphid clusters, and the fine network further locates aphid individuals. The average detection accuracy is 76.8% on their test set. Pei et al. (2020) introduced a candidate region generation method based on ORB (Oriented FAST and Rotated Brief) (Rublee et al., 2011) features, which can quickly identify suspected aphid regions, and incorporated it into a CNN-based detection framework for the detection of aphids. Their experimental results showed that using the proposed method for generating candidate regions in the detection network can improve the average detection precision by 0.281. Li et al. (2022) proposed a density map estimation network based on MCNN (Zhang et al., 2016) and FPN (Lin et al., 2017), and applied it to the aphid counting task. Through the experimental verification on their test set, they achieved MAE (Mean Absolute Error) and MSE (Mean Square Error) of 10.22 and 12.24, respectively. Additionally, they conducted comparison experiments under different scenarios with varying density distributions of aphids. The results showed that the counting accuracy based on the detection boxes is higher under low-density distribution of aphids, while the counting accuracy based on the density map estimation is higher under high-density distribution of aphids.

Overall, the aphid counting method based on image processing is suitable for simple, ideal and single-species scenarios, it is difficult to adapt to more complex situations. As it typically employs threshold segmentation or other typical segmentation methods based on digital image processing techniques in the first step to extract regions of interest from the images. However, these typical segmentation methods are highly sensitive to changes in lighting, color, and other factors, making it difficult to adapt to complex aphid counting scenarios. The aphid counting method based on traditional machine learning heavily relies on prior knowledge for effective hand-crafted feature designing. By contrast, the aphid counting method based on deep learning is more accurate and robust. Moreover, the counting based on the detection network has advantages in scenarios where the aphids are sparsely distributed, as it predicts every possible aphid using a bounding box. But it tends to miss detections when the distribution positions of different aphid individuals are connected or partially overlapped, or when the aphids are extremely small. On the other hand, the counting based on the density map estimation network employs regression to predict the probability that each pixel may represent an aphid, which can better deal with the cases where there are connections and partial overlaps between different aphid individuals. But it sometimes mistakes the background for the real target, leading to over-prediction, especially when the background is extremely complex.

The two of the biggest challenges in counting aphids are that aphids are small objects (1 to 10 millimeters) and that their density distribution varies across different areas of the field. These challenges have greatly increased the difficulty of automatic aphid counting. To this end, we proposed a hybrid network architecture integrating the detection network and the density map estimation network for an accurate aphid counting in sugar beet fields. It can autonomously determine which branch to use for better counting estimation. If the aphid density in the image is low, it utilizes the detection network to accurately count aphids. Otherwise, it switches to the density map estimation network to accurately estimate the number of aphids. In this paper, we will only focus on accurate aphid detection and counting without being able to differentiate the specific aphid species.

## 2. Materials and methods

*2.1 Dataset and pre-processing*

The aphid dataset was collected by the BBRO (British Beet Research Organisation) using smartphones in multiple sugar beet fields in the UK in 2020. The original dataset consists of high-resolution images. Training models with high-resolution images is time-consuming and requires a substantial amount of GPU memory. Moreover, the original dataset contains many invalid regions that do not contain aphids, leading to an imbalance between positive and negative samples. Therefore, we preprocessed the original dataset. Firstly, we cut the image into a square area. Then, we divided the square area into four equal parts and removed images that don't contain aphids. The final aphid dataset contains 412 images.

We used LabelImg[1] and Labelme[2] to label aphids, and these ground truths were confirmed by an experienced entomologist from BBRO. Initially, all images were labelled with bounding box labels using LabelImg, which are used for training of the detection network. Afterwards, point labels were added to the images using Labelme, based on the bounding box labels generated by LabelImg. And we used the Gaussian kernel function to convolve the coordinates of points representing the position of aphids to obtain the corresponding continuous density map. The generated density map is used to train the density map estimation network. The complete preprocessing and labeling flow of the aphid dataset is shown in Fig. 1.

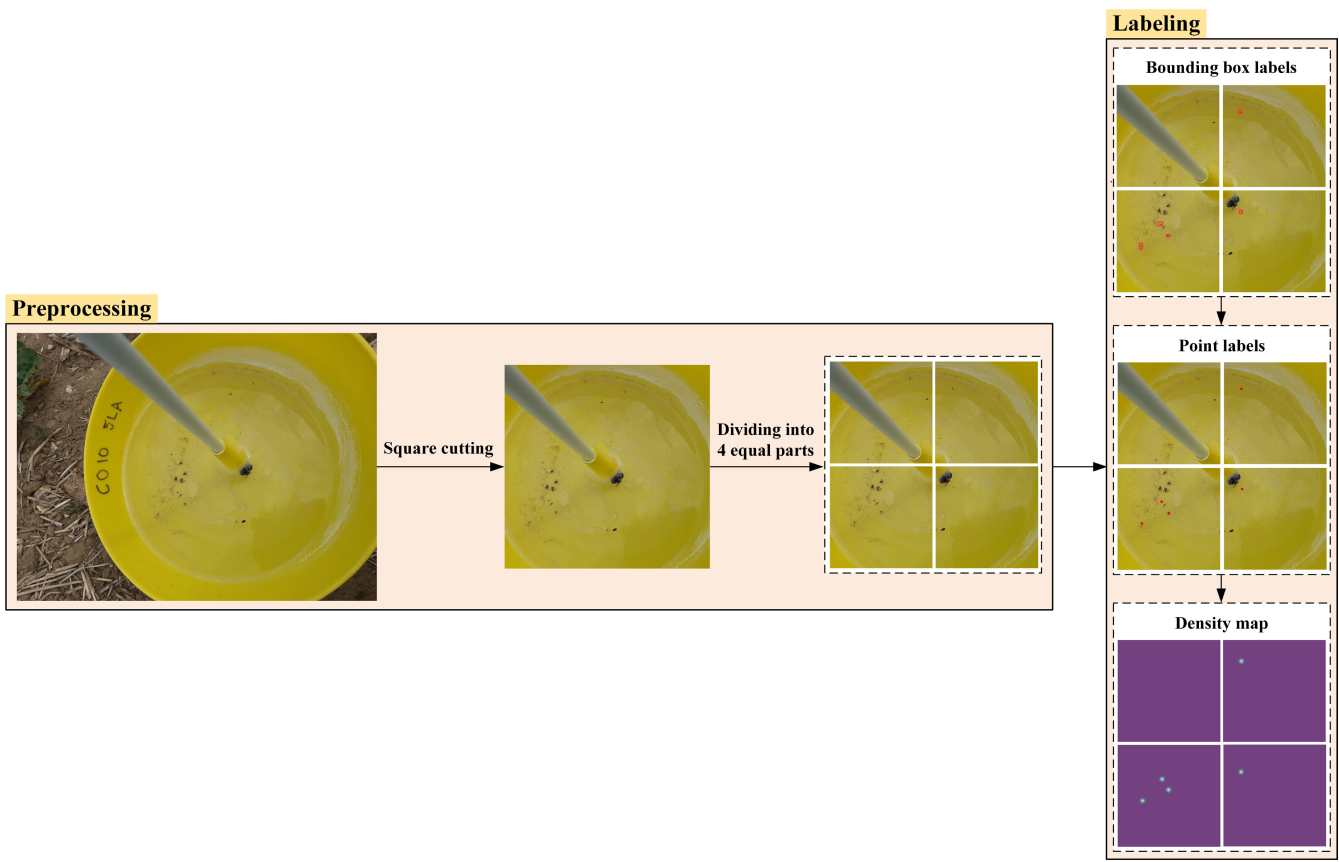

---

[1]Available in: https://github.com/heartexlabs/labelImg

[2]Available in: https://github.com/wkentaro/labelme

**Fig. 1.** The complete preprocessing and labeling flow of the aphid dataset.

As the original input image is typically scaled to a square size before being input to the CNN, using square cutting can ensure that the aspect ratio of bounding boxes of aphids in the input of CNN is consistent with that of the original image.

According to the literature survey (Li et al., 2022), we found that counting based on the density map estimation network outperforms counting based on the detection networks in the case of high-density distribution of aphids. To leverage this advantage, we screened images with high-density distribution of aphids from the original aphid dataset of 412 images to form a separate dataset for training the density map estimation network model. The dataset with high-density distribution of aphids contains 63 images. Overall, there are two aphid datasets: Aphid_ Dataset_1 (Standard aphid dataset) and Aphid_ Dataset_2 (High-density aphid dataset). A summary and comparison of two aphid datasets is shown in Table 1.

**Table 1** The summary and comparison of two aphid datasets.

| Name | Number of images | Aphid distribution density level | Division ratio (training : validation : test) |
| --- | --- | --- | --- |
| Aphid_ Dataset_1 | 412 | normal | 8:1:1 |
| Aphid_ Dataset_2 | 63 | high | 2:1:1 |

Since Aphid_ Dataset_2 is relatively small, we adopt the concept of 2-fold Cross Validation (Zhou, 2021) and divide it into training, validation and test setsin a ratio of 2:1:1 instead of the commonly used 8:1:1.

*2.2 Dataset analysis*

In order to further understand the characteristics of the aphid dataset and the difficulties of automatic aphid counting, we made data statistics and analysis on the standard aphid dataset Aphid_ Dataset_1. Fig. 2 shows the histogram of the number of aphids in Aphid_ Dataset_1. Fig. 3 shows a proportion chart for the number of aphid bounding boxes of different sizes. The metrics for classifying aphids as small, medium and large refer to the thresholds used in the COCO dataset (Lin et al., 2014). The examples of aphid images with annotated bounding boxes in Aphid_ Dataset_1 are displayed in Fig. 4.

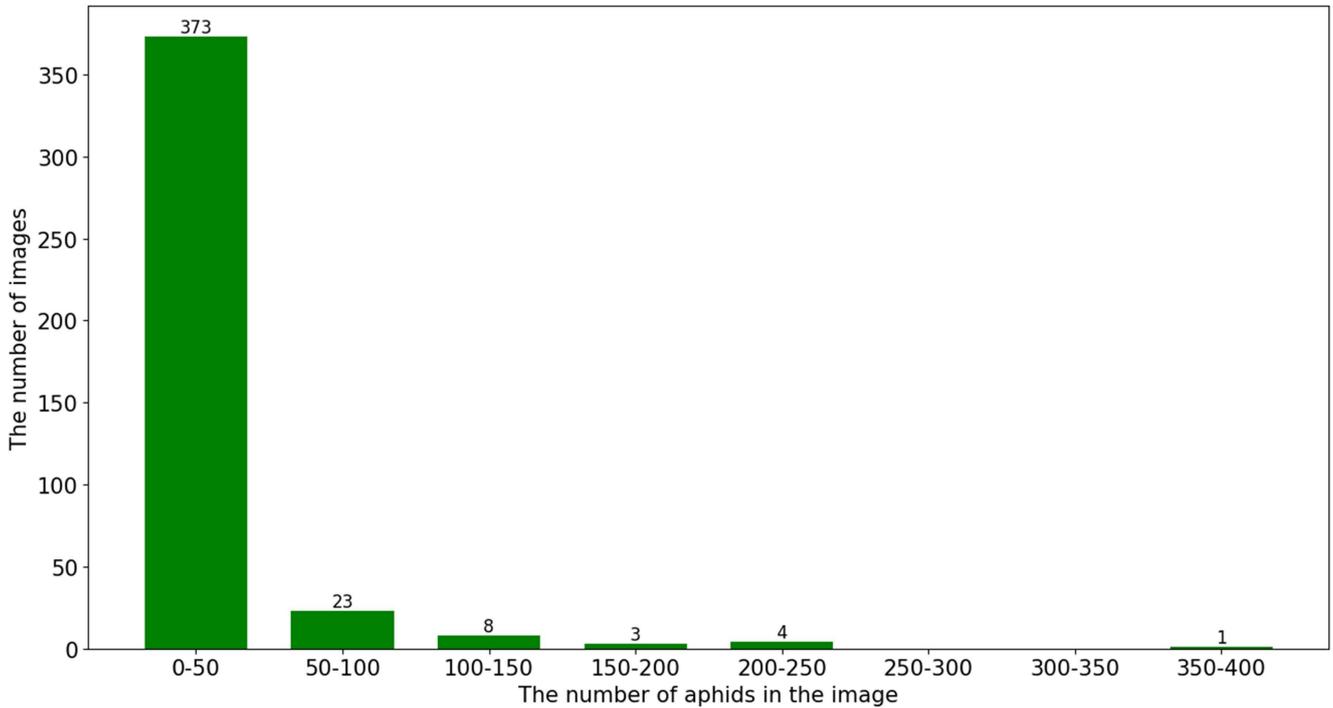

**Fig. 2.** The histogram of the number of aphids in Aphid_ Dataset_1.

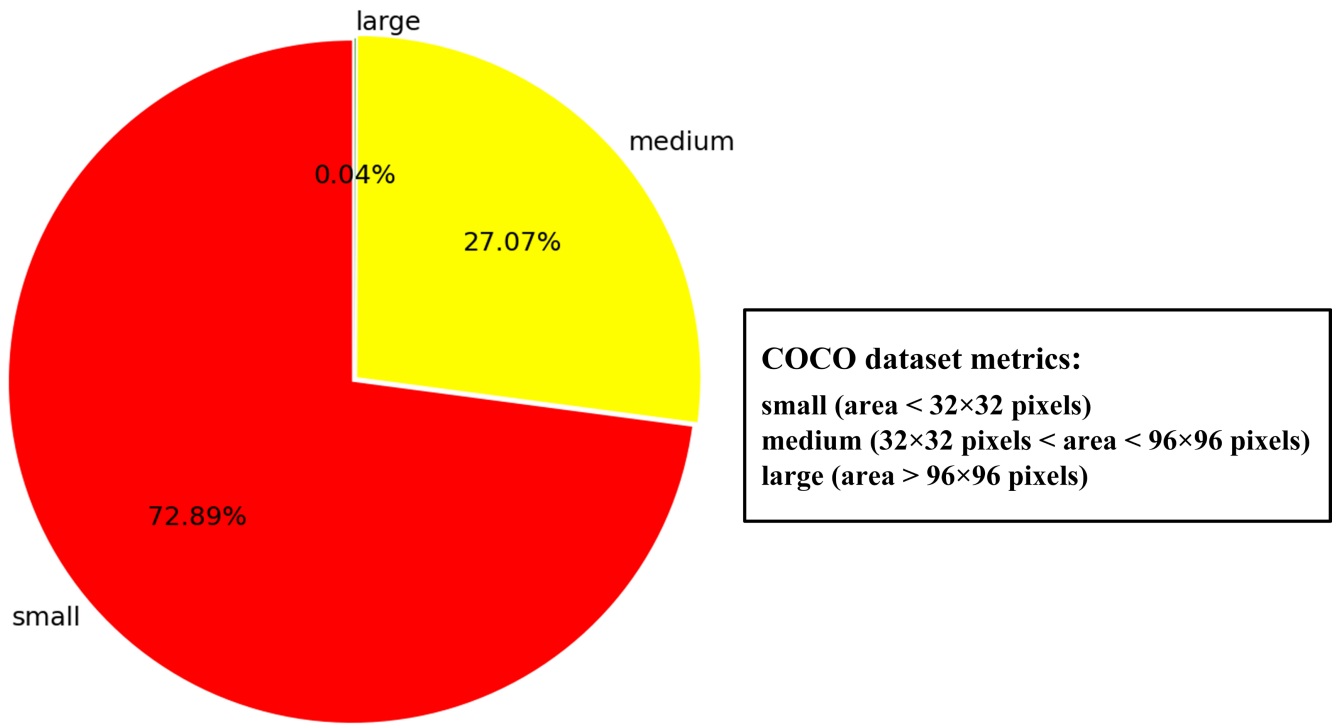

**Fig. 3.** The proportion chart of the number of aphid bounding boxes in different sizes (Lin et al., 2014).

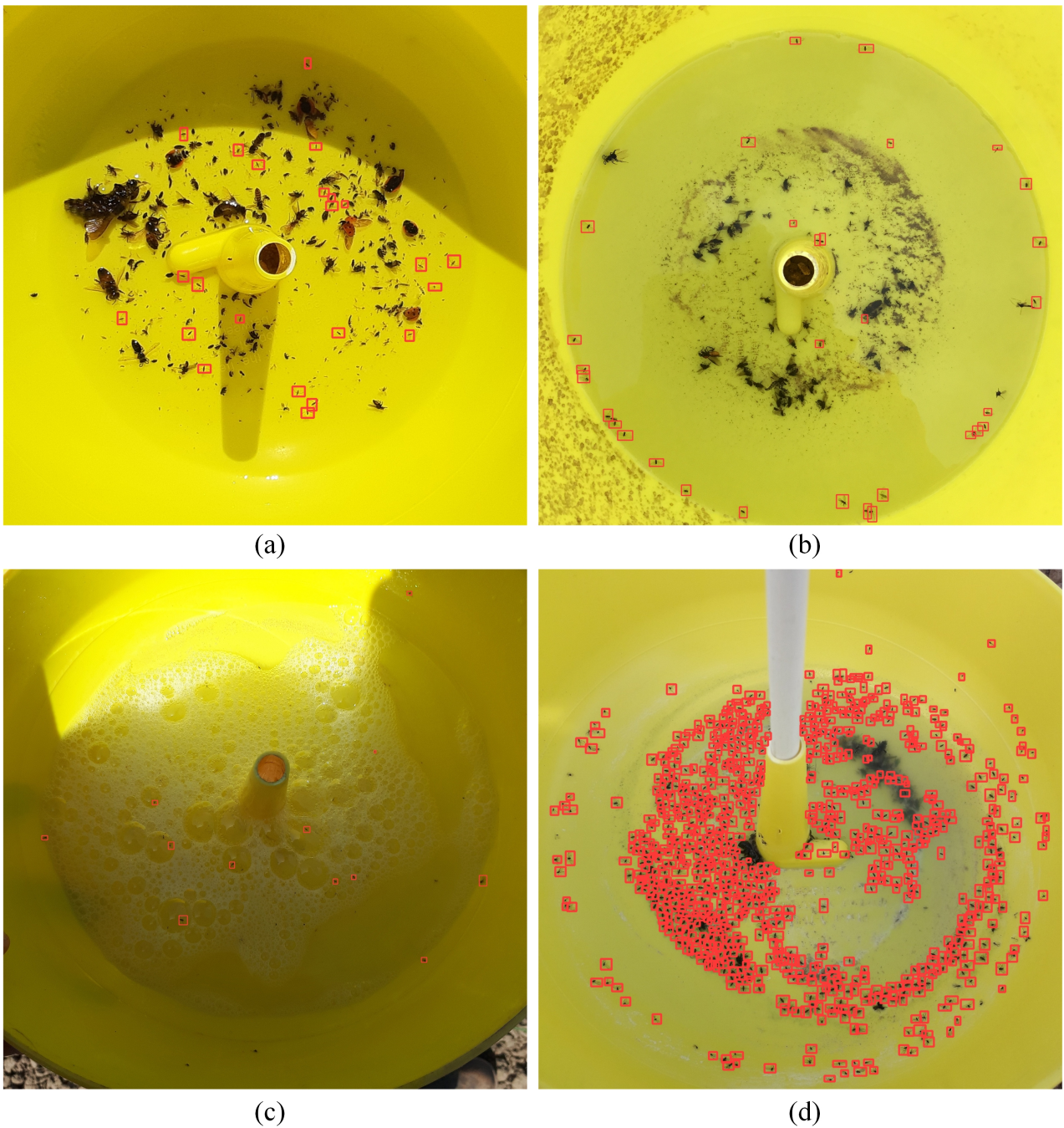

**Fig. 4.** Some typical samples of aphid images with annotated bounding boxes.

It can be seen from Fig. 2 that the number of aphids in most of images are less than 50. A small proportion of the images show a high-density distribution of aphids, where the numbers of aphids in individual images exceeds 200.

Fig. 3 illustrates that the majority of aphids in the standard aphid dataset (72.89%) belong to the small object category, while a very small proportion (0.04%) of aphids are categorized as large objects.

Fig. 4 more intuitively shows the distribution characteristics of aphids in the yellow water pan trap. From Fig. 4(a) and Fig. 4(b), it can be seen that in addition to the aphids we are interested in, it also contains many other types of pests or insects, such as mosquitoes, ladybugs, bees and so on. It is particularly important to note that some pests are very similar to aphids in morphology, which can easily lead to confusion in the counting task. At the same time, aphid individuals differ slightly in size and appearance as they are in different growth stages. Moreover, there are some soil debris distributed in dispersion or aggregation in the yellow water pan trap due to the natural environment of fields. Fig. 4(c) shows that aphids look blurred when they are submerged in the water with bubbles. And their posture often varies greatly due to the buoyancy of the water. Fig. 4(d) shows that when aphids are densely distributed in the yellow water pan trap, different aphid individuals are prone to connect and partially overlap.

To sum up, the standard aphid dataset has the following characteristics: 1) The density distribution of aphids in the dataset varies, with most samples showing a low density of aphids and only a few samples showing a high density of aphids. 2) The majority of aphids are small objects, and therefore the aphid counting can be considered as a task of counting small objects. 3) The background of aphid counting scenarios is complex. And there are phenomena of interclass similarity and intraclass variation between aphids and others pests. 4) When aphids are submerged in the water, their posture will change variously along with three dimensions. 5) Serious connections and partial overlaps occur when aphids are densely distributed in the yellow water pan trap.

*2.3 The proposed automatic aphid counting network architecture*

*2.3.1 Overview of our proposed automatic aphid counting network*

In this paper, we proposed a hybrid network architecture for automatic aphid counting, which integrates the detection network and the density map estimation network. When the aphid density in the image is low, the detection network is employed to count aphids. To make the detection network better suited for the aphid counting task, we proposed an improved Yolov5 (Jocher et al., 2021) as the detection network. Conversely, when the aphid density in the image is high, the density map estimation network CSRNet (Li et al., 2018) is used for counting aphids. An overview of our proposed automatic aphid counting network architecture is shown in Fig. 5.

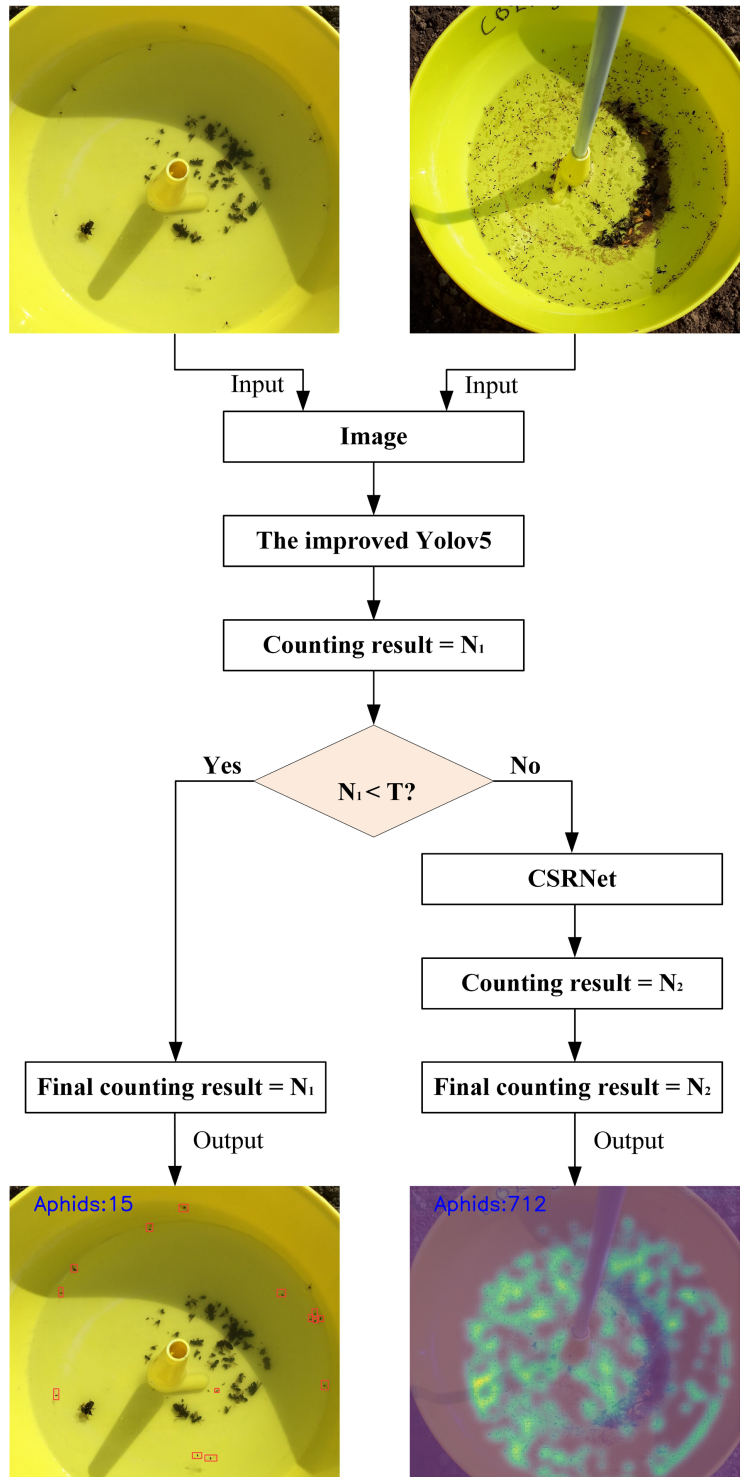

**Fig. 5.** Overview of our proposed automatic aphid counting network architecture.

As we can see from Fig. 5, it first inputs the image into the improved Yolov5 to detect and count the number of aphids, which is represented as $N_1$. If $N_1$ is less than the pre-defined threshold T, it means the distribution density of aphids in the image is low. The final counting result is considered to be $N_1$.

On the contrary, if $N_1$ is greater than or equal to T, it means the distribution density of aphids in the image is high. It switches the counting network from the improved Yolov5 to CSRNet, and get the final counting result as $N_2$. How to determine the optimal T is explained in detail in Section 3.6.

*2.3.2 The improved Yolov5*

As a popular detection network framework, Yolov5 achieves a balanced trade-off between detection speed and accuracy by incorporating the grid regression concept from single-stage networks (SSD (Liu et al., 2016), RetinaNet (Lin et al., 2017)) for end-to-end detection, and introducing the anchor mechanism from two-stage networks (Faster R-CNN (Ren et al., 2015), Cascade R-CNN (Cai and Vasconcelos, 2018)) to enhance precision in target localization. Then, certain design principles in Yolov5 confer advantages in the detection of small objects (such as aphids): a) Yolov5 introduces a novel anchor box method employing the K-means clustering algorithm to automatically and precisely determine anchor box sizes, enabling better adaptability to various target dimensions. This implies that Yolov5 can more effectively learn and accommodate features of small objects during the training process. b) Yolov5 incorporates the FPN + PAN (Liu et al., 2018) structure, enhancing the network's capability to capture targets of varying scales. This design is particularly advantageous for small object detection. c) Yolov5 employs various data augmentation techniques, notably mosaic augmentation, by concatenating four images as the final training data. This increases the quantity of small objects in the training dataset, ensuring effective learning of small objects. Besides, the network complexity of Yolov5 can be customized to make it easier to adapt to different detection tasks. Based on different network widths and depths, Yolov5 can be categorized into five versions: Yolov5n, Yolov5s, Yolov5m, Yolov5l, and Yolov5x. Considering the current limitations in computing resources and our future plan to deploy the automatic aphid counting model to edge devices, this paper adopts Yolov5s, which has relatively low complexity but also ensures higher detection accuracy, as the foundational framework. We also took note of the latest YOLO series, Yolov7 (Wang et al., 2023) and Yolov8 (Jocher, 2023). Considering that the aphid dataset used in this paper is relatively small, and aphids are considered small objects, we have decided to stick with Yolov5 as the foundational framework. This is because Yolov7 (415 layers) is a deeper network compared to Yolov5s (270 layers). On a relatively small dataset, a deeper network like Yolov7 leads to redundant parameters, hindering the network from better learning target features, and making it susceptible to gradient vanishing issues. Additionally, Yolov5 employs the anchor-based mechanism, while Yolov8 utilizes the anchor-free mechanism. Anchor-based methods, with predefined anchor boxes, offer better handling of localization ambiguity and scale variation, contributing to higher detection accuracy compared to anchor-free methods. Finally, given the characteristics of the standard aphid dataset and the challenges of counting aphids summarised in Section 2.2, we proposed an improved Yolov5 to better adapt the aphid counting task. The pipeline of the improved Yolov5 is shown in Fig. 6.

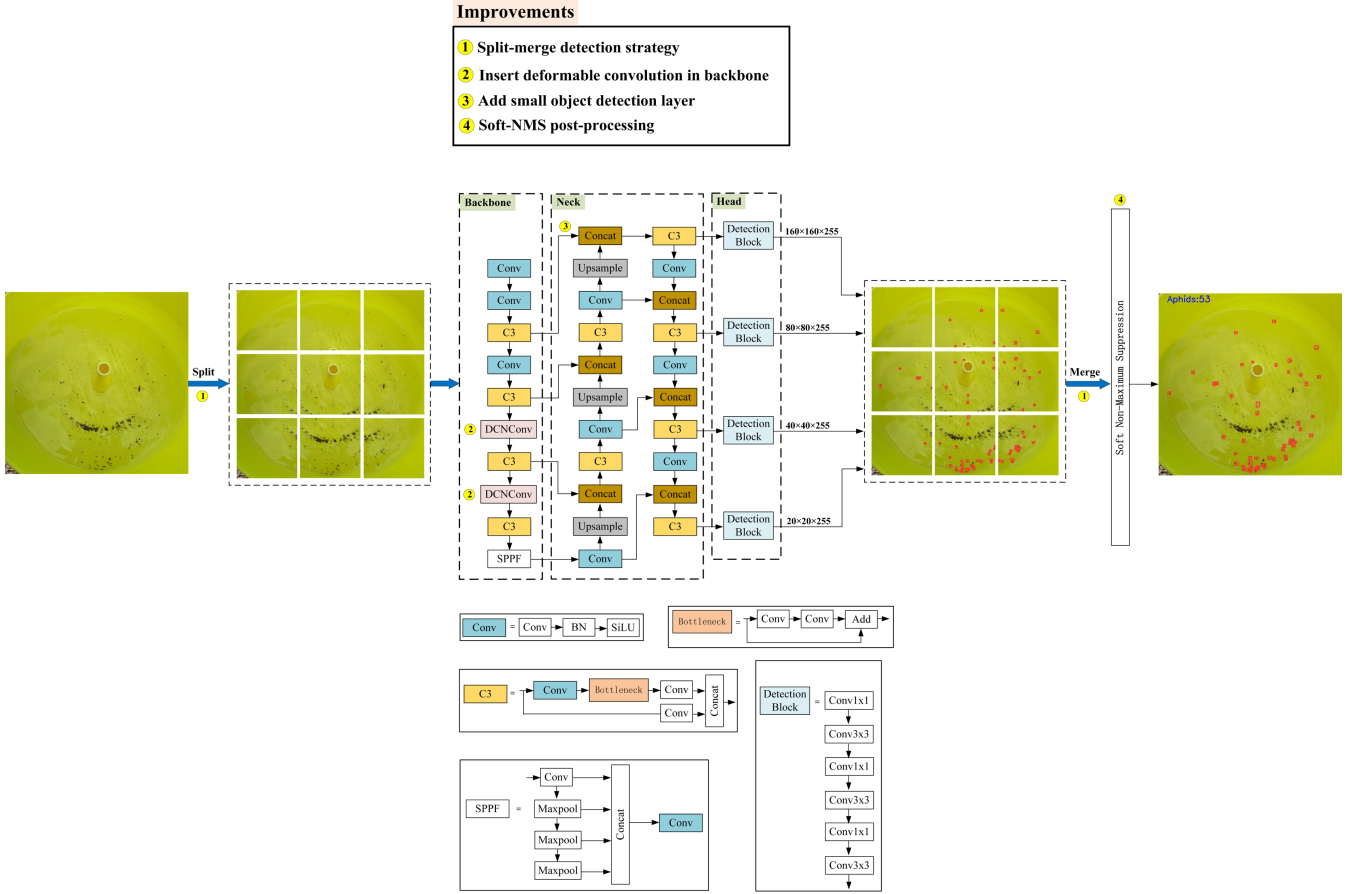

**Fig. 6.** The pipeline of the improved Yolov5.

The improved Yolov5 has four main improvements over the original Yolov5:

a) Split-merge detection strategy

Being inspired by (Van Etten, 2018) and (Akyon et al., 2022), we use a split-merge detection strategy to improve the network's performance in detecting small objects. Firstly, as seen in Fig. 6, we split the input image into equal sized blocks using a sliding window (256 × 256). To prevent some aphids distributed at the boundary of different blocks, from being missed whilst splitting the images, we set the overlapping rate between adjacent blocks to 20%. It ensures that the whole body of aphid can be seen at least once. Then, it conducts model inference on each image block and obtains multiple corresponding block detection results. Finally, these block detection results are merged and a post-processing pipeline is developed to produce the final detection result in an image.

b) Insert deformable convolution in backbone

The posture of aphids often varies in the yellow water pan trap due to the buoyancy of water. To adapt the network to this change, we introduce the deformable convolution (Dai et al., 2017) in the backbone of the improved Yolov5. Some Conv (Conv+BN+SiLU) modules are replaced with the DCNConv (Deformable Convolutional Networks) modules. The standard convolution uses convolution kernels with a fixed size (such as $5 \times 5$, $3 \times 3$, $1 \times 1$), so the receptive field of the same CNN layer is fixed. In reality, different objects have different shapes, the shape of the same object will change when it is viewed from different perspectives, and the scale of the same object will vary at different positions. It is difficult for the standard convolution to deal with these changes. However, the deformable convolution adds a 2D offset (x,y) to each convolution sampling point, allowing the receptive field to adapt to the shape of objects. Moreover, the offset is learnable. Therefore, the deformable convolution is more effective in handling geometric transformations caused by variable factors, such as changes in perspective, scale, posture, and partial deformation.

Since the offset of the deformable convolution is learned through gradient backpropagation across additional convolution units, it unavoidably leads to an increase in computational workload for the network. Furthermore, the deformable convolution may introduce extraneous contextual information that could potentially interfere with the normal process of feature extraction. Therefore, we only replace the last two Conv modules in the backbone with DCNConv modules instead of replacing all of them.

c) Add small object detection layer

The sizes of the last feature maps in Yolov5 are $80 \times 80$, $40 \times 40$ and $20 \times 20$, respectively. Considering that the standard size of the input image of Yolov5 is $640 \times 640$, we can calculate that the sizes of the corresponding detection object in the input image are 8 × 8, 16 × 16 and 32 × 32, respectively. As a result, if the width or height of an object in the input image is less than 8 pixels, it will be difficult for Yolov5 to detect. However, aphids are usually very small objects. To improve the performance of Yolov5 in detecting small objects, we add a small object detection layer to the network. We can see from Fig. 6 that it continues to upsample to generate the 160 × 160 feature maps after layer 17, and concatenates them with the feature maps from layer 2. Compared with the original Yolov5, it adds 160 × 160 feature maps in the head of the network, in addition to the 80 × 80, 40 × 40 and 20 × 20 feature maps. Overall, by incorporating larger feature maps, it becomes possible to detect very small objects in an image.

d) Soft-NMS post-processing

Compared with the original Yolov5, the improved Yolov5 uses soft-NMS post-processing (Bodla et al., 2017) instead of NMS (Non Maximum Suppression) (Neubeck and Van Gool, 2006).

With regard to the NMS in the original Yolov5, it first finds the candidate box $A$ with the highest confidence. Then it calculates the IOU (Intersection Over Union) area of all remaining candidate boxes $B = \{b_0, b_1, ... b_n\}$ to $A$. Finally, the score $S_i$ of the candidate boxes with IOU greater than the threshold $T$ is set to 0. This means that the candidate boxes with IOU greater than the threshold $T$ are deleted. It is formulated as in Equation (1). As the NMS only considers the IOU ratio between candidate boxes, if $T$ is set too low, some candidate boxes will be erroneously deleted. Conversely, if $T$ is set too high, it will lead to an increase in false detections.

$$S_i = \begin{cases} S_i, & \text{IoU}(A, b_i) < T \\ 0, & \text{IoU}(A, b_i) \geq T \end{cases} \tag{1}$$

The soft-NMS, an improved NMS, takes into account both the scores and the overlap ratio of the candidate boxes, so it has better performance than the NMS. As shown in Equation (2), it uses weights to update the score $S_i$ of candidate boxes when the IOU is greater than the threshold $T$. If the IOU is higher, $S_i$ will become smaller. In contrast, if the IOU is lower, $S_i$ will become larger. The ultimate goal is to retain those candidate boxes with low IOU but high confidence.

$$s_i = \begin{cases} S_i, & \text{IoU}(A,b_i) < T \\ S_i(1-\text{IoU}(A,b_i)), & \text{IoU}(A,b_i) \geq T \end{cases} \quad (2)$$

*2.3.3 The introduction of CSRNet*

CSRNet mainly is composed of a front-end network and a back-end network. The front-end network consists of the first 10 convolution layers and 3 pooling layers of VGG-16 (Simonyan and Zisserman, 2014), which is used to extract features of the area of interest in the image. The back-end network is composed of six convolution layers with a dilation rate of 2 that can expand the receptive field without altering the size of feature maps. Finally, a 1 ×1 ordinary convolutional layer is used to output the predicted density map. Since the front-end network uses three pooling layers with $2 \times 2$, the output size of the density map in the last layer is 1/8 of the original input. To ensure that the final output result has the same resolution as the input image, it employs bilinear interpolation with a factor of 8 to scale the output of the density map and then obtain the final predicted density map. The structure diagram of CSRNet is shown in Fig. 7.

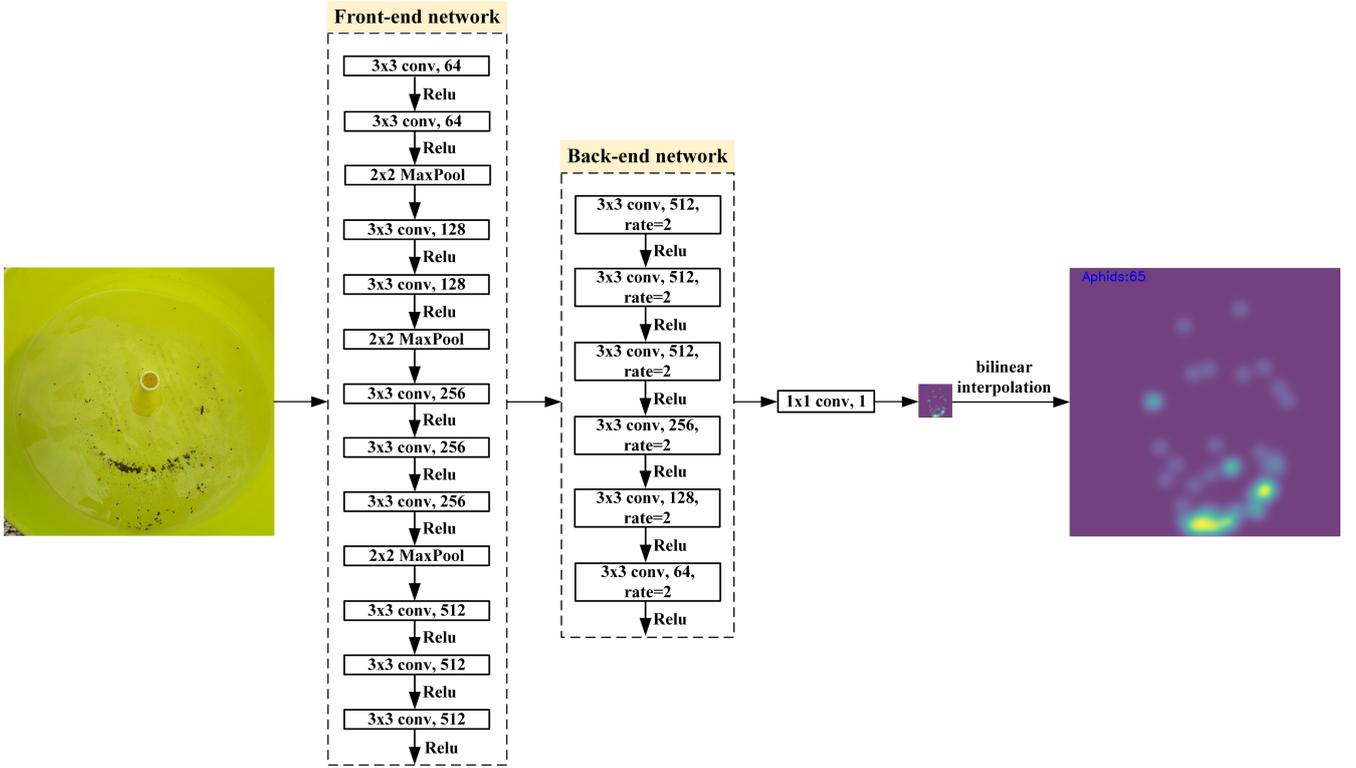

**Fig. 7.** Structure diagram of CSRNet.

*2.4 Evaluation metrics*

*2.4.1 Evaluation metrics of the detection model*

We use AP (Average Precision) to evaluate the accuracy of the detection model. The calculation of AP is shown in Equation (3)-(5).

$$P = \frac{TP}{TP + FP} \tag{3}$$

$$R = \frac{TP}{TP + FN} \tag{4}$$

$$AP = \int_0^1 P(R)d(R) \tag{5}$$

In the above equations, TP (True Positives) indicates the number of positive samples that are correctly detected as positive classes. FP (False Positives) is the number of negative samples falsely classified as positive classes. FN (False Negatives) is the number of positive samples falsely classified as negative classes. P (Precision rate) refers to the correct proportion of all targets detected. R (Recall rate) refers to the correct proportion in all positive samples. P and R are usually a pair of contradictory measures. Conventionally, using P and R alone may not accurately reflect the detection performance. To obtain a more comprehensive measure, we use AP as a metric for the accuracy of the detection model. The AP calculates the average precision value for the recall value over 0 to 1, which means the area under the precision-recall curve. As the AP takes into account both P and R, it is more suitable for evaluating detection accuracy of the model. In general, the higher the test AP of a model, the better the detection performance.

*2.4.2 Evaluation metrics of the density map estimation model*

We use MAE and RMSE (Root Mean Squared Error) to evaluate the accuracy and robustness of the density map estimation model. The calculation of MAE and RMSE is shown in Equation (6) and Equation (7) respectively.

$$MAE = \frac{1}{N}\sum_{1}^{N}\left|y_i - y_i^{'}\right| \tag{6}$$

$$RMSE = \sqrt{\frac{1}{N}\sum_{1}^{N}(y_i - y_i^{'})^2} \tag{7}$$

In the above equations, $y_i$ is the predicted number of aphids obtained by integrating the predicted density map of the image $i$. $y_i^{'}$ is the actual number of aphids in the image $i$. The MAE is the mean value of absolute error, which can reflect the error of the predicted value. The RMSE is the square root of the ratio of the square sum of the deviation between the observed value and the true value to the number of observations $N$, which is used to measure the deviation between the observed value and the true value. The smaller the MAE and RMSE, the higher the counting accuracy of the model.

In addition, since the counting method based on the density map estimation network can not calculate the AP, we use MAE and RMSE as the unified evaluation metrics when comparing the accuracy of counting aphids between the detection model and the density map estimation model.

*2.5 Implementation Details*

Our proposed automatic aphid counting network was implemented on the deep learning framework PyTorch. The training and testing equipment is an NVIDIA GeForce GTX 1060 GPU with 6 GB memory. We trained the models using the original Yolov5, the improved Yolov5, CSRNet and our proposed automatic aphid counting network on Aphid_ Dataset_1 and Aphid_ Dataset_2 respectively.

For the model training of detection networks including the original Yolov5 and the improved Yolov5, we used partial pre-trained weights from the COCO dataset to expedite convergence and reduce training time. The training parameter settings of detection networks are as shown Table 2. It is important to note that the default number of training epochs in Yolov5 is set to 100, and the early stopping is employed to obtain the optimal model. In order to ensure that our model thoroughly learns the features of the data, we have set the number of training epochs to 600. To obtain the best model, we used a default fitness function from Yolov5, which is a weighted combination of AP@0.5 and AP@0.5:0.95 with the former contributing 10% of the weight and the latter contributing the remaining 90%. It is important to note that we did not use an early stopping mechanism for model training. This is because we aimed to compare the training processes of the original Yolov5 and the improved Yolov5 (as shown in Fig. 8). The convergence speeds of the original Yolov5 and the improved Yolov5 differed during the training process, making it difficult to choose a consistent early stopping parameter "patience" that would apply equally well to both models. Choosing not to use early stopping mechanism ensured that the comparison between different models (the original Yolov5 and the improved Yolov5) was more comparable, as they were trained under the same conditions without any interruptions. Furthermore, we believed that the early stopping mechanism in the original Yolov5 is somewhat unreasonable because it only monitored the mAP value (stopping training if the mAP did not increase for a certain number of consecutive epochs) without simultaneously considering the change in loss value. In such cases, if the early stopping parameter "patience" was not set properly, it might lead to prematurely terminate training, even when the loss was still decreasing. This could result in missing opportunities for further performance improvement of the models, as the models may not have truly converged at that point. Additionally, to prevent overfitting, the dataset was augmented during training with random translation, random scaling, random left-right flip, color augmentation, and mosaic augmentation.

**Table 2** The training parameter settings of detection networks.

| Parameter | Configuration |
| --- | --- |
| Input size | $640 \times 640$ |
| Batch size | 4 |
| Optimizer | SGD (Stochastic Gradient Descent) |
| Initial learning rate | 0.01 |
| Decay | 0.0005 |
| Momentum | 0.937 |
| Loss | BCEWithLogitsLoss with FocalLoss |
| Training epochs | 600 |

For the model training of CSRNet, the 10 convolutional layers of the front-end network used the pre-trained weights of ImageNet (Deng et al., 2009) for initialization and fine-tuning training, and the parameters of other convolution layers were initialized using the Gaussian distribution with the standard deviation of 0.01. The training parameter settings of CSRNet are as shown Table 3. It's worth noting that the network employs random cropping to extract sub-images, constituting 1/4 the size of the original images, as input. This approach can be regarded as a form of data augmentation. Additionally, the default setting for training epochs in CSRNet is 400. Similarly, to ensure comprehensive learning of data features, we have extended the number of training epochs to 2000. During the training process, the model selection is based on comparing the value of MAE. The model with the lowest MAE on the validation set is considered the best model.

Table 3 The training parameter settings of CSRNet.

| Parameter | Configuration |
| --- | --- |
| Input size | 1/4 of the original image size |
| Batch size | 1 |
| Optimizer | SGD (Stochastic Gradient Descent) |
| Initial learning rate | 1e-7 |
| Decay | 0.0005 |
| Momentum | 0.95 |
| Loss | MSELoss |
| Training epochs | 2000 |

# 3. Results

*3.1 Training results*

  To intuitively understand the convergence status of the model during the training process and assess whether the training parameters are set appropriately, visualizations were conducted for the training loss, validation loss, and validation accuracy of the detection networks, including the original Yolov5, our improved Yolov5, and the density map estimation network CSRNet, on the standard aphid dataset Aphid_Dataset_1. The training loss curve, validation loss curve and mAP curve of the original Yolov5 and our improved Yolov5 are shown in Fig. 8. The loss curve consists of three parts: cls_loss (classification loss), box_loss (bounding box regression loss), obj_loss (objectness loss).The training loss curve, validation loss curve, MAE curve and RMSE curve of CSRNet are shown in Fig. 9.

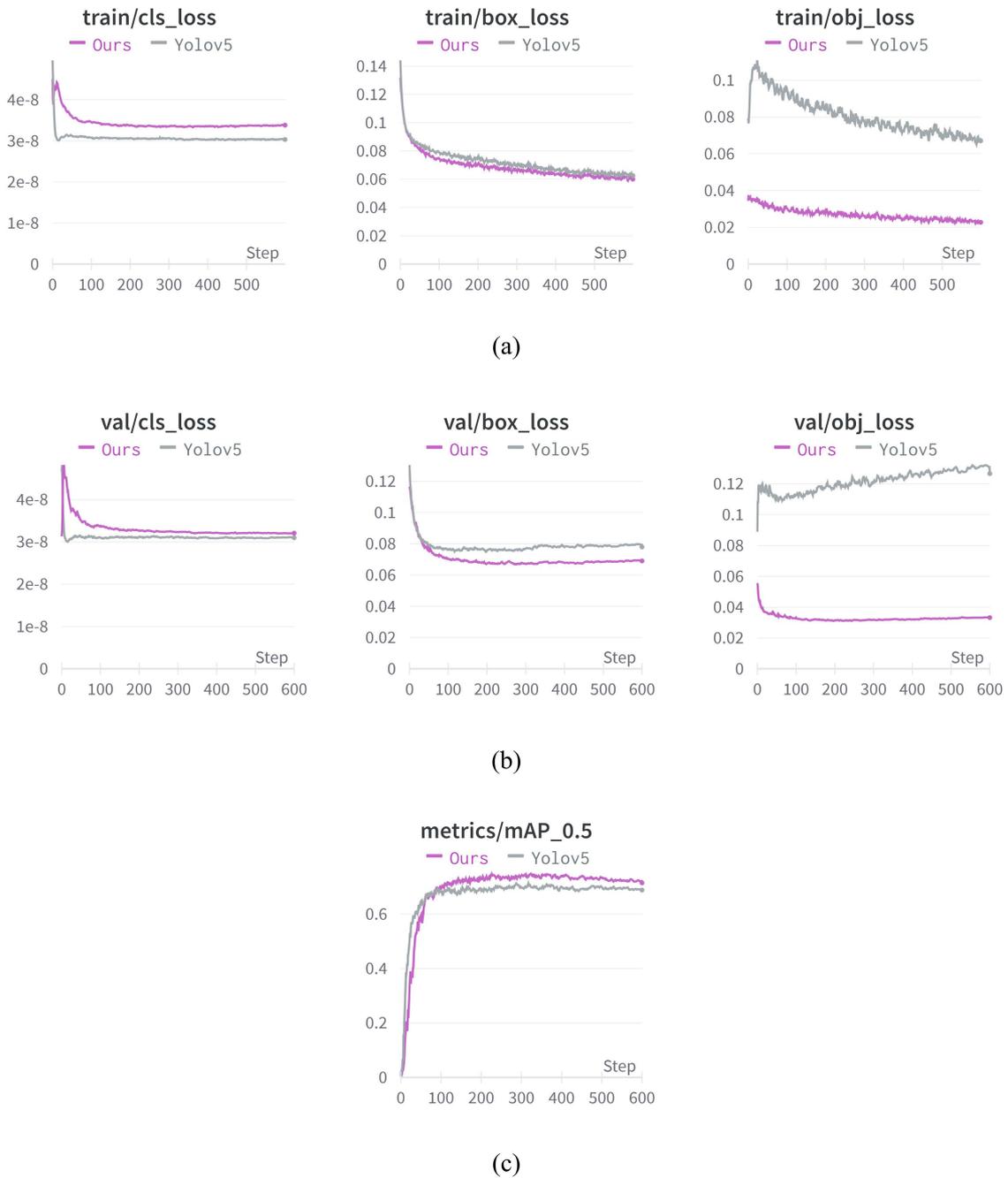

**Fig. 8.** Visualization of training results - Loss and accuracy for detection networks on the Aphid_Dataset_1. (a) Training loss curve. (b) Validation loss curve. (c) mAP curve on validation set.

It can be seen from Fig. 8 that the training loss curve, validation loss curve of our improved Yolov5 are relatively normal and stable during the whole of training process. The training loss and validation loss drop rapidly in the first 100 epochs. After that, both of them are gradually leveling off. They reach the convergence state after about 300 epochs. However, the serious overfitting occurs in the original Yolov5. After about 70 epochs, the training loss is still in a rapid decline stage, but the validation loss has started to rise rapidly. In addition, we can also note that the training loss curve, validation loss curve of our improved Yolov5 are lower than that of the original Yolov5. And after about 100 epochs, the mAP of our improved Yolov5 on validation set apparently higher than that of the original Yolov5. This indicates that our improved Yolov5 is unquestionably superior to the original Yolov5.

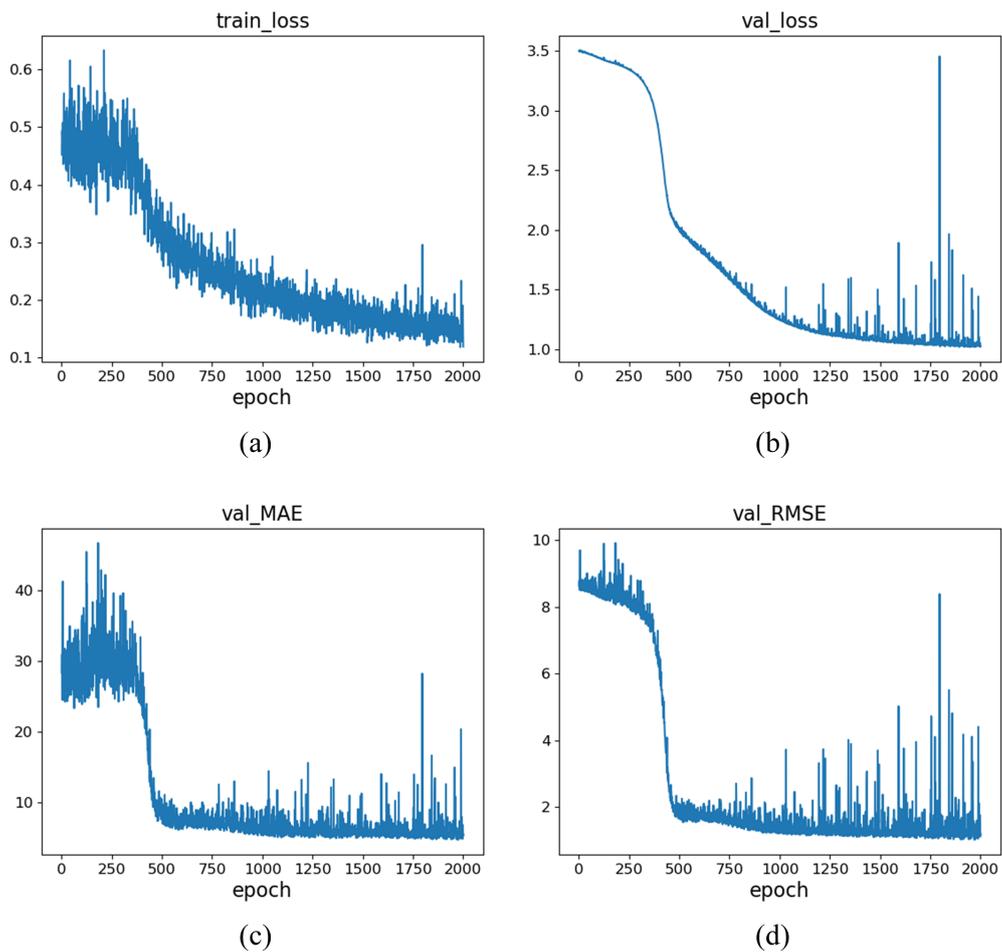

**Fig. 9.** Visualization of training results - Loss and accuracy for CSRNet on the Aphid_Dataset_1. (a) Training loss curve. (b) Validation loss curve. (c) MAE curve on validation set. (d) RMSE curve on validation set.

From Fig. 9, it can be observed that for the first 1000 epochs, the training and validation loss curves of CSRNet experience a rapid descent followed by a gradual decline, ultimately reaching a converged state around 1500 epochs. The validation MAE and RMSE curves exhibit a similar pattern, with a swift decrease in the initial 500 epochs, followed by a more gradual descent, and stabilizing approximately after 1500 epochs.

In summary, from Fig. 8 and Fig. 9, it can be observed that the training parameters specified in Table 2 and Table 3 are reasonable. This is because, within the predefined training parameters, the models of both the improved Yolov5 and CSRNet have reached a stable convergence state without exhibiting overfitting or underfitting phenomena. It is also verified that there is no need to set an early stopping mechanism for the training of both the improved Yolov5 and CSRNet models. But it is worth noting that during the training process of the original Yolov5, overfitting occurred. After about 70 epochs, while the training loss continued to decline rapidly, the validation loss started to rise rapidly. However, the mAP was still in an upward trend (Fig. 8). Therefore, even if an early stopping mechanism was set during training of the original Yolov5, it would not be effective. This is because, as mentioned earlier, the early stopping mechanism in the original Yolov5 is somewhat unreasonable because it only monitored the mAP value (stopping training if the mAP did not increase for a certain number of consecutive epochs) without simultaneously considering the change in loss value. Additionally, to ensure comparability between the original Yolov5 and the improved Yolov5, there was no need to set an early stopping mechanism for training of the original Yolov5.

*3.2 Evaluation on the test set*

To evaluate our proposed automatic aphid counting network, we conducted a comprehensive set of experiments that consists of three parts. Firstly, we compared the aphid counting performance of our proposed network against several other networks on the test set of the standard aphid dataset Aphid_Dataset_1. The comparison results are shown in Fig. 10.

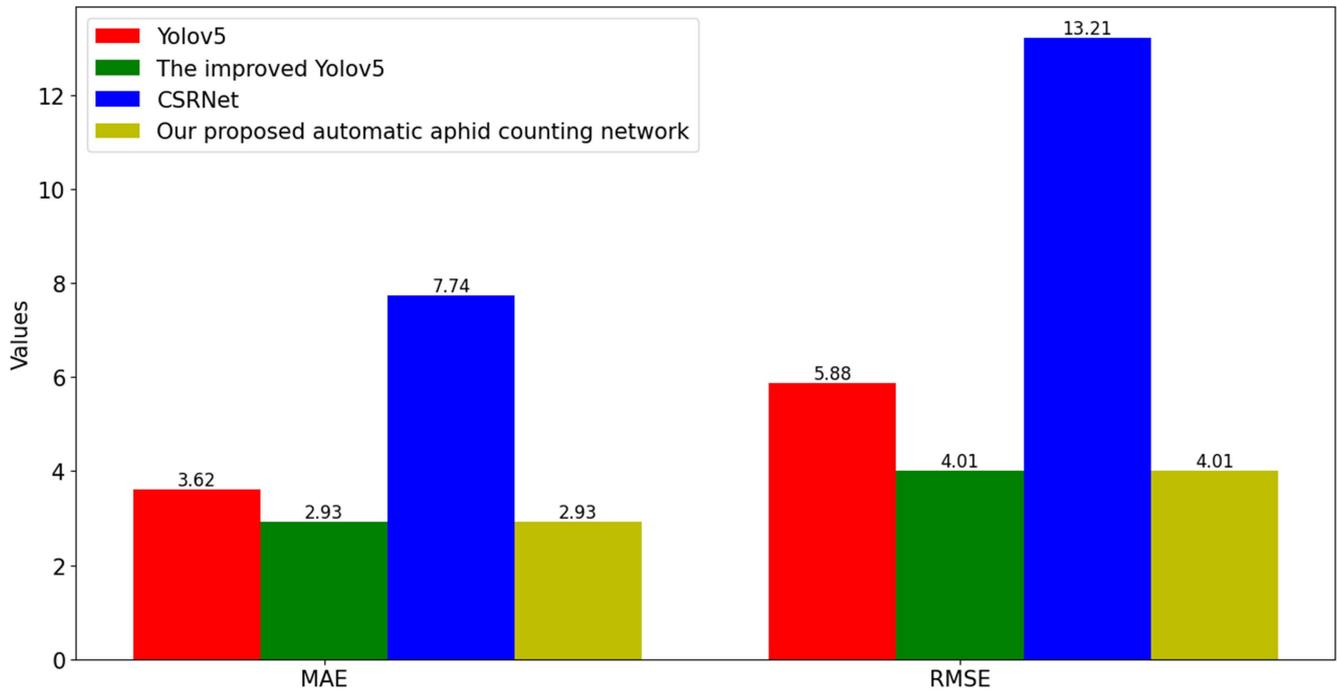

**Fig. 10.** The comparison results of counting aphids using different networks on the test set of Aphid_ Dataset_1.

As shown in Fig. 10, our proposed automatic aphid counting network, as well as the improved Yolov5, individually achieve the lowest MAE value (2.93) and the lowest RMSE value (4.01) compared with other counting networks. The reason why the above two networks achieve the same MAE and RMSE is because the images on the test set of Aphid_Dataset_1 are considered as low density of aphids. On the other hand, it is worth noting that the MAE and RMSE of both the original Yolov5 and the improved Yolov5 are significantly smaller than those of CSRNet. This is also because the images on the test set of Aphid_Dataset_1 exhibit low density of aphids. Thus, the superior performance of CSRNet in counting densely distributed objects was not demonstrated on the test set of Aphid_Dataset_1 compared to the detection networks. In our upcoming experiment, we would further investigate the aphid counting performance of different networks in Aphid_Dataset_2 with the high-density aphid distribution.

Secondly, we evaluated the aphid detection performance of the improved Yolov5 and compared it against the original Yolov5 on the test set of the standard aphid dataset Aphid_Dataset_1. Additionally, we compared the aphid detection performance of detectors based on the Yolov5 framework with several detectors, including classical detectors (Faster R-CNN, Cascade R-CNN, RetinaNet) and advanced detectors (Yolov7, Yolov8). The backbones of Faster R-CNN, Cascade R-CNN, and RetinaNet all use ResNet50 (He et al., 2016). To ensure the credibility of the comparative experiments, all compared models were subjected to a uniform experimental environment, and training parameters were maintained consistently in accordance with the original settings. The comparison results are shown in Table 4.

Table 4 The comparison results of detecting aphids using different detection networks on the test set of Aphid_ Dataset_1.

| Method | AP (Average Precision) |
| --- | --- |
| Faster R-CNN | 0.20 |
| Cascade R-CNN | 0.16 |
| RetinaNet | 0.14 |
| Yolov5(s) | 0.60 |
| Yolov7 | 0.16 |
| Yolov8 (s) | 0.59 |
| The improved Yolov5 (s) | 0.65 |

As shown in Table 4, firstly, the test results of classical detectors, Faster R-CNN, Cascade R-CNN, and RetinaNet on the test set of Aphid_Dataset_1, all have an AP of less than or equal to 20%, which is quite poor. Secondly, detectors from the YOLO series achieved relatively better test results, except for Yolov7, which had a testing AP of only 16%. Additionally, the original Yolov5 exhibited a testing AP greater than Yolov7 and Yolov8, reaching 60%. This indicates that choosing Yolov5 as the foundational framework for this paper is reasonable, as explained in Section 2.3.2. Finally, the improved Yolov5 achieves a 5% higher AP than the original Yolov5. This result is in line with our first experiment shown in Fig. 10, where the MAE and RMSE of the improved Yolov5 are smaller than those of the original Yolov5.

Finally, we investigated the aphid counting performance of different networks in scenarios where aphids are densely distributed. To accomplish this, we tested several different networks mentioned above on the high-density aphid dataset Aphid_Dataset_2 and compared their counting results. The comparison results are shown in Fig. 11.

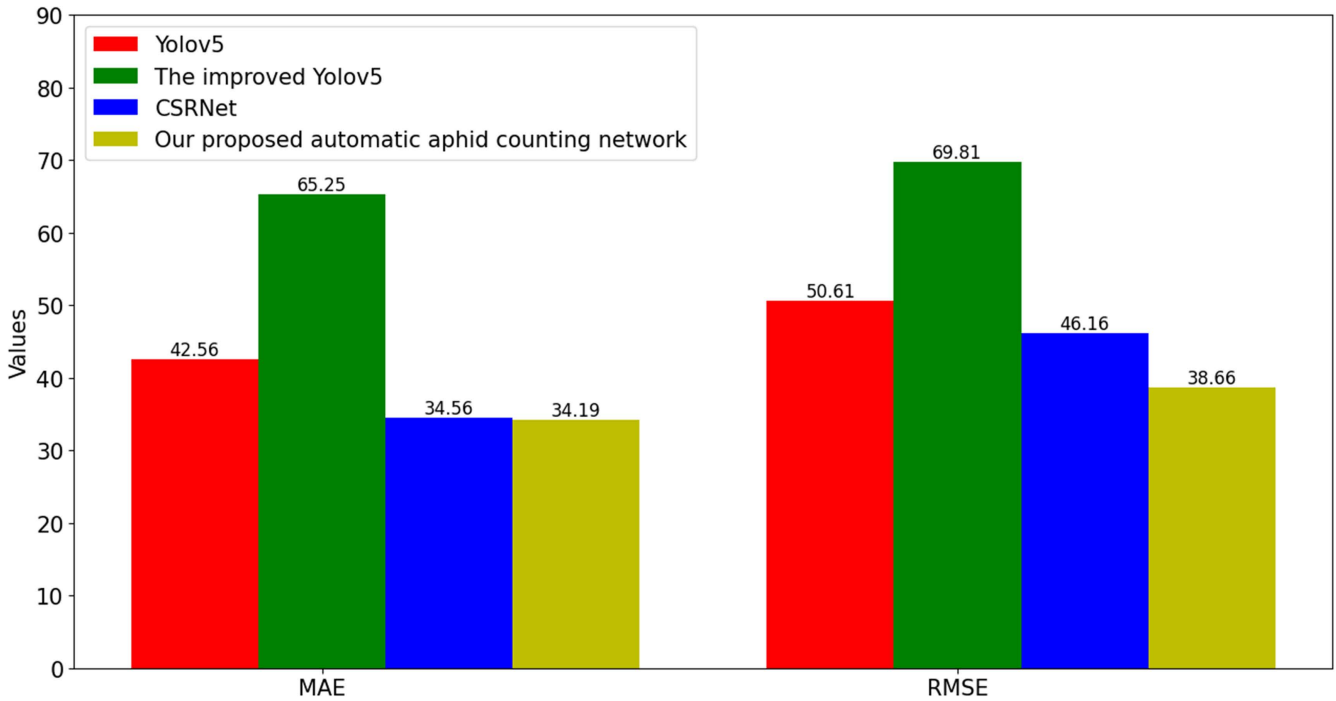

**Fig. 11.** The comparison results of counting aphids using different networks on the test set of Aphid_ Dataset_2.

As shown in Fig. 11, our proposed automatic aphid counting network achieves the lowest MAE

(34.19) and RMSE (38.66) when evaluating on the high-density aphid dataset Aphid_Dataset_2. The most noteworthy point is that the MAE and RMSE of CSRNet are significantly lower than those of both the original Yolov5 and the improved Yolov5, with values of 34.56 and 46.16, respectively. It highlights the superior performance of CSRNet for counting aphids when aphids are densely distributed, surpassing that of both the original Yolov5 and the improved Yolov5. Additionally, the MAE and RMSE of the improved Yolov5 are greater than those of the original Yolov5. This discrepancy can be attributed to the high density of aphids that are present in the Aphid_Dataset_2. When we use the split-merge detection strategy in the improved Yolov5, many complete aphids located on the boundary of different blocks are often cut into different parts and distributed to different blocks, resulting in these aphids not being detected due to low classification confidence.

Overall, it can be concluded that when the distribution density of aphids is standard, the counting performance based on the detection network is better than that based on the density map estimation network. On the contrary, when the distribution density of aphids is high, the counting performance based on the density map estimation network outperforms that based on the detection network. These findings support the rationale behind our proposed automatic aphid counting network, which integrates the detection network and the density map estimation network to achieve the best possible performance across a range of different aphid distribution conditions in real-field aphid counting tasks. In addition, the improved Yolov5 demonstrated a superior aphid counting performance compared with the original Yolov5 in the standard aphid dataset.

*3.3 Visualization of aphid counting results*

To evaluate the generalization of our proposed automatic aphid counting network and to make a more intuitive comparison of aphid counting results across different methods, we randomly selected five sample images with varying aphid distribution densities that are not part of the original aphid dataset, and then used models trained by different networks to count aphids in these images. The results of counting aphids using different networks are shown in Fig. 12. The first column shows the original images with the ground truth bounding boxes, while the second, third, fourth, and fifth columns display the counting results obtained using the original Yolov5, the improved Yolov5, CSRNet, and our proposed automatic aphid counting network, respectively. At the same time, for a deeper understanding of the prediction results, we separately calculated the TP, FN, and FP values for each test image's corresponding predicted results. We visualized the results using confusion matrices, as shown in Fig. 13. It is worth noting that we computed and displayed confusion matrices only for the original Yolov5 and the improved Yolov5, without calculating one for CSRNet. This is because the density map estimation network uses regression to predict density values at each pixel in the image, which are continuous values rather than discrete values. Therefore, it can not be used for confusion matrix calculation.

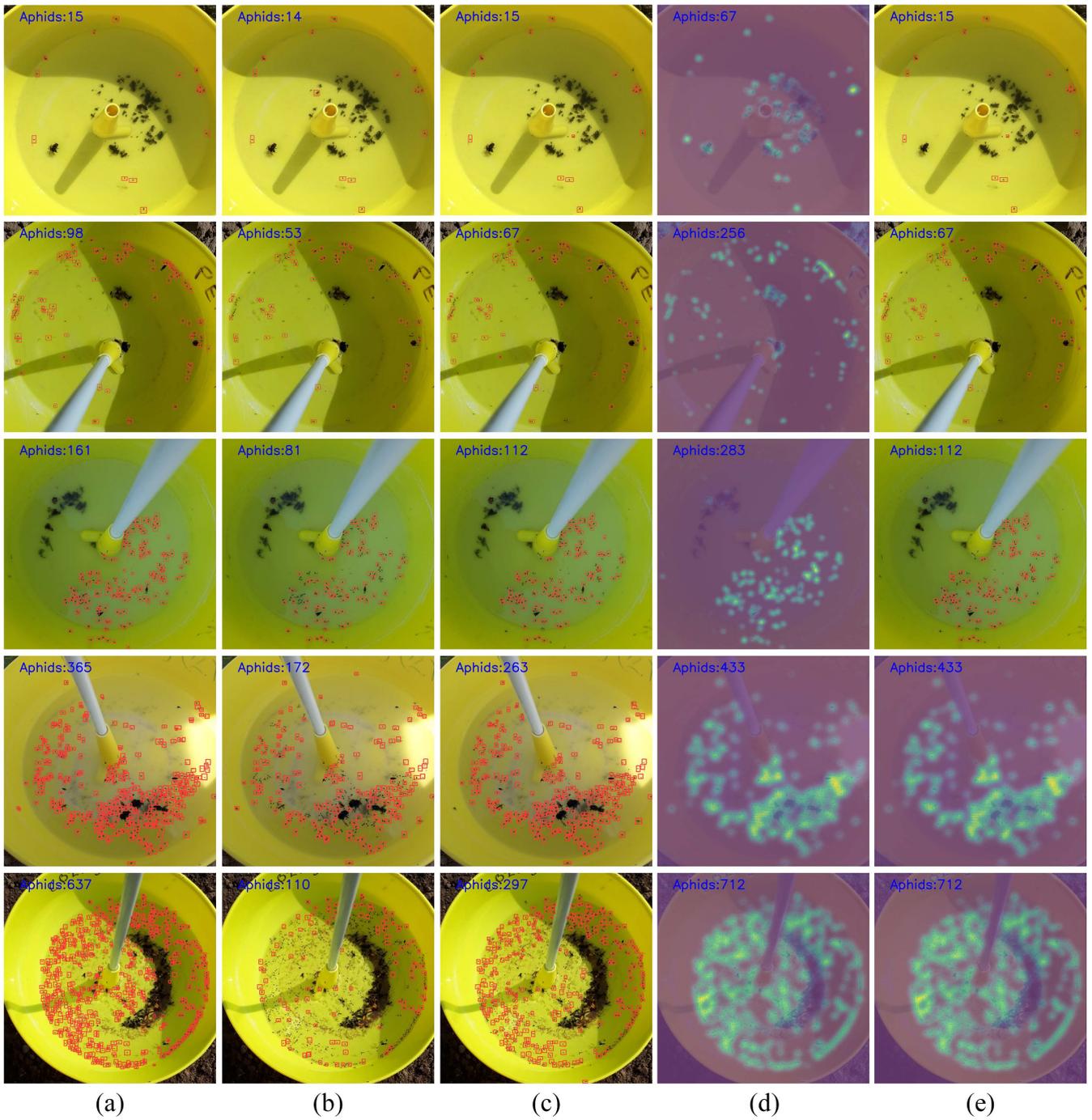

**Fig. 12.** Visualisation of aphid counting results using different networks on various sample images with varying aphid distribution densities. (a) Original images with the ground truth bounding boxes. (b) The counting results by Yolov5. (c) The counting results by the improved Yolov5. (d) The counting results by CSRNet. (e) The counting results by our proposed automatic aphid counting network.

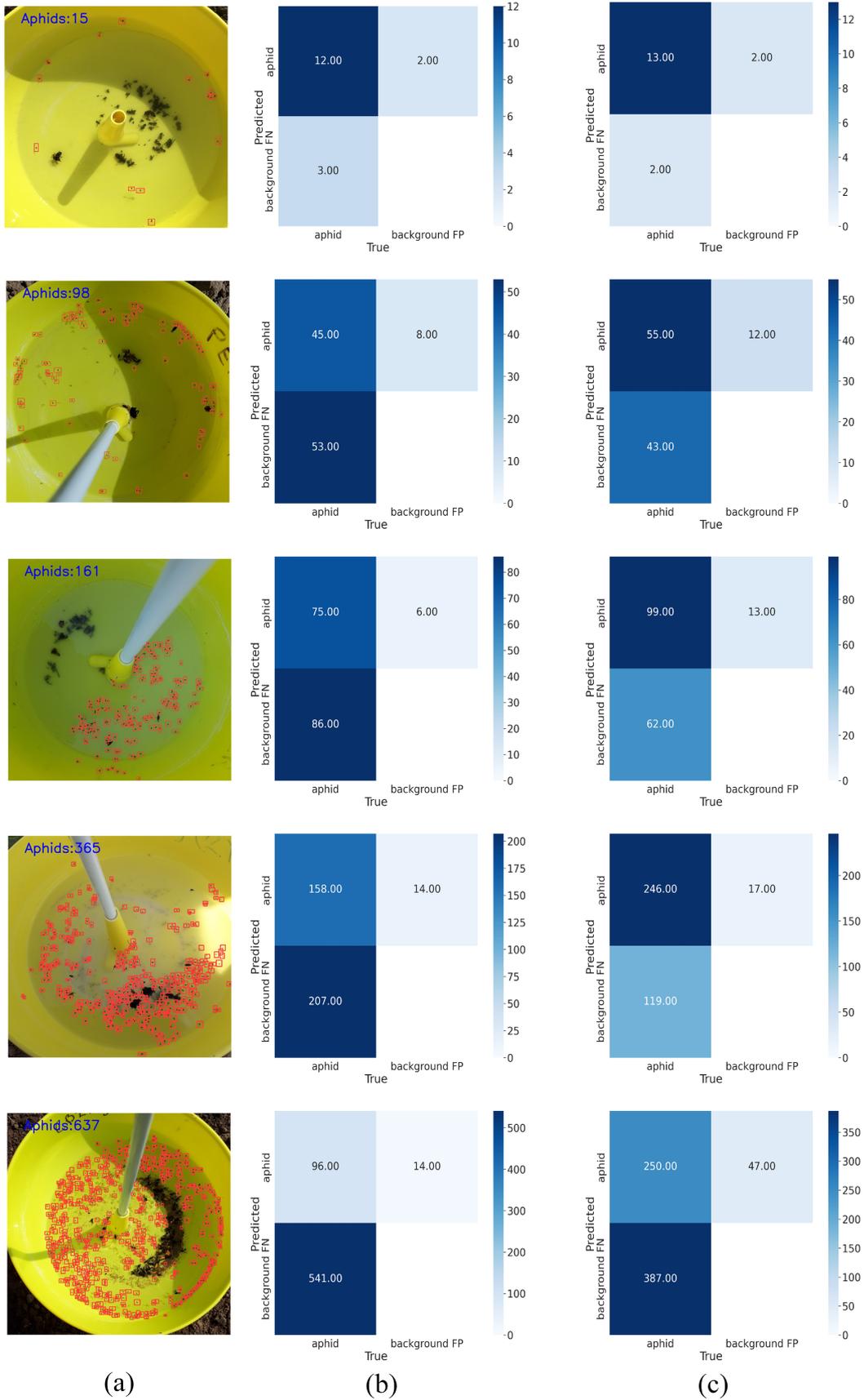

(a)             (b)             (c)

**Fig. 13.** Confusion matrices of predicted results corresponded to Fig. 12. (a) Original images with the ground truth bounding

boxes. (b) Confusion matrix of the results predicted by Yolov5. (c) Confusion matrix of the results predicted by the improved Yolov5.

It can be seen from Fig. 12 that the original Yolov5 and the improved Yolov5 have comparable counting performance when the distribution density of aphids is very low. However, as the distribution density of aphids increases, the improved Yolov5 outperforms the original Yolov5 in accurately counting aphids. Nevertheless, both methods exhibit a gradual decline in counting accuracy as the distribution density of aphids increases. The higher the distribution density of aphids, the more pronounced the decline in counting accuracy. At the same time, the counting effect of the improved Yolov5 is much better than that of the original Yolov5 in local areas where aphids are densely distributed or very small in shape. In contrast, the counting results of CSRNet exhibit a notable over-prediction when the distribution density of aphids is low. However, as the distribution density of aphids increases, the degree of over-prediction gradually decreases. In fact, when the distribution density of aphids is very high, CSRNet achieves excellent aphid counting results, outperforming the detection-based methods by a significant margin.

From Fig. 13, we can observe that firstly, the TP values in the predicted results of the improved Yolov5 are consistently greater than the TP values in the predicted results of the original Yolov5. Especially as the distribution density of aphids increases, the difference between them becomes more significant. Therefore, it can be concluded that the prediction performance of the improved Yolov5 is superior to the original Yolov5, particularly when the distribution density of aphids increases, this superiority becomes more evident. Secondly, whether in the predicted results of the original Yolov5 or the improved Yolov5, the FN values are significantly larger than the FP values. This implies that the model fails to predict the number of aphids far more often than it incorrectly predicts the background region as aphids. Moreover, as the distribution density of aphids increases, the FN values noticeably and continually increase. In extreme cases, when the distribution density of aphids is extremely high, the FN values even surpass the TP values. Consequently, when using the detection networks to count aphids, the issue of missing aphids is quite severe.

*3.4 Ablation experiments*

The improved Yolov5 mainly makes four improvements to the original Yolov5: a) Insert deformable convolution in the backbone, b) Add small object detection layer, c) Split-merge detection strategy, d) Soft-NMS post-processing. a) and b) are the improvements to the network itself, while c) and d) are the improvements of the pre-processing and post-processing techniques. To further investigate the effect of each improvement component on the aphid detection, we conducted ablation experiments on the test set of standard aphid dataset Aphid_Dataset_1. The ablation experiments are divided into two groups. The first group focused on the effect of the network itself, which includes a) and b). The second group analysed the effect of pre-processing and post-processing techniques on the basis of the first group, which includes c) and d). The settings of the training parameters for the ablation experiments are the same as those described in Section 2.5. The results of the ablation experiments are presented in Table 5 and Table 6.

**Table 5** Results of ablation experiments on the network itself of the improved Yolov5 with Aphid_ Dataset_1.

| Yolov5s | Deformable convolution | Small object layer | Split-merge | Soft-NMS | AP (Average Precision) |
|---|---|---|---|---|---|
| √ |   |   |   |   | 0.60 |
| √ | √ |   |   |   | 0.59 |
| √ |   | √ |   |   | 0.62 |

| Yolov5s | Deformable convolution | Small object layer | Split-merge | Soft-NMS | AP (Average Precision) |
|---------|------------------------|--------------------|-------------|----------|-----------------------|
| √ | √ | √ | | | 0.64 |

Table 6 Results of ablation experiments on the pre-processing and post-processing techniques of the improved Yolov5 with Aphid_ Dataset_1.

| Yolov5s | Deformable convolution | Small object layer | Split-merge | Soft-NMS | AP (Average Precision) |
|---------|------------------------|--------------------|-------------|----------|-----------------------|
| √ | √ | √ | | | 0.64 |
| √ | √ | √ | √ | | 0.63 |
| √ | √ | √ | | √ | 0.64 |
| √ | √ | √ | √ | √ | 0.65 |

Table 5 shows that when we only insert the deformable convolution into the backbone of the network, the AP on the test set decreases by 1%. However, when we add the small object detection layer to the network, the AP on the test set increases by 2%. Notably, when we combine the deformable convolution with the small object detection layer in the network, the AP on the test set increases by 4%. It demonstrates that using these two components together yields better improvement in detection accuracy compared with using them separately.

Table 6 reveals that using split-merge detection strategy or soft-NMS alone does not significantly affect the AP on the test set. However, when we combine these two components, the AP on the test set increases by 1%.

Given that the aphid dataset used in this paper is relatively small. To assess whether the learning capability of the improved Yolov5 can be enhanced with a larger set of images, we augmented the standard aphid dataset Aphid_Dataset_1 by including an additional 12 training images, forming an augmented dataset called Aug_Aphid_Dataset_1. Subsequently, we conducted model training, followed by ablation experiments as shown in Table 5 and Table 6. The outcomes of these ablation experiments are detailed in Table 7 and Table 8.

**Table 7** Results of ablation experiments on the network itself of the improved Yolov5 with Aug_Aphid_ Dataset_1.

| Yolov5s | Deformable convolution | Small object layer | Split-merge | Soft-NMS | AP (Average Precision) |
|---|---|---|---|---|---|
| √ | | | | | 0.63 |
| √ | √ | | | | 0.64 |
| √ | | √ | | | 0.62 |
| √ | √ | √ | | | 0.65 |

**Table 8** Results of ablation experiments on the pre-processing and post-processing techniques of the improved Yolov5 with Aug_Aphid_ Dataset_1.

| Yolov5s | Deformable convolution | Small object layer | Split-merge | Soft-NMS | AP (Average Precision) |
|---|---|---|---|---|---|
| √ | √ | √ | | | 0.65 |
| √ | √ | √ | √ | | 0.63 |
| √ | √ | √ | | √ | 0.65 |
| √ | √ | √ | √ | √ | 0.64 |

Comparing the test results between Table 5 and Table 7 reveals that, when focusing on the network itself of the improved Yolov5, the testing mAP of the model trained on the augmented dataset (Aug_Aphid_Dataset_1) exhibits an overall improvement compared to the model trained on the standard aphid dataset (Aphid_Dataset_1). Subsequently, a comparison of the test results in Table 6 and Table 8, focusing on the pre-processing and post-processing techniques of the improved Yolov5, indicates that the testing mAP of the model trained on the augmented dataset does not show improvement over the model trained on the standard aphid dataset. Overall, expanding the aphid dataset can enhance the testing accuracy of the model to a certain extent.

*3.5 Setting of the size of the sliding window and the overlapping area ratio*

When we use the split-merge detection strategy in the improved Yolov5, we need to set the size of the sliding window and the overlapping area ratio. To investigate the optimal values of the size of the sliding window and the overlapping area ratio for the standard aphid dataset used in this paper, we conducted the test experiments. As the input image is uniformly resized to $640 \times 640$ before being input to the network, we set the size of the sliding window to $32 \times 32$, $64 \times 64$, $128 \times 128$, $256 \times 256$, and $512 \times 512$, respectively, and conducted aphid detection test on the test set of the standard aphid dataset Aphid_Dataset_1. The experimental results are shown in Table 9. Similarly, we set the overlapping area ratio from 0 to 0.5 at an interval of 0.1 and conducted aphid detection test on the test set of the standard aphid dataset Aphid_Dataset_1. The experimental results are shown in Table 10.

**Table 9** The aphid detection results with different sizes of the sliding window on the test set of Aphid_ Dataset_1.

| Size of the sliding window | AP (Average Precision) |
| --- | --- |
| $32 \times 32$ | 0.38 |
| $64 \times 64$ | 0.57 |
| $128 \times 128$ | 0.61 |
| $256 \times 256$ | 0.65 |
| $512 \times 512$ | 0.63 |

Table 9 shows that AP increases as the size of the sliding window increases. This is because when the sliding window is too small, the image will be cut into many blocks, causing complete aphids located on the boundary of different blocks to be split into different parts and distributed to different blocks. These aphids may not be detected easily due to low classification confidence. It is worth noting that the maximum AP of 0.65 is achieved when the sliding window is set to $256 \times 256$. However, as the size of the sliding window further increases to 512 × 512, the AP decreases. This is because a larger sliding window makes it difficult for the network to capture the features of small objects, such as aphids.

Table 10 The aphid detection results with different overlapping area ratios on the test set of Aphid_ Dataset_1.

| Overlapping area ratio | AP (Average Precision) |
|---|---|
| 0 | 0.62 |
| 0.1 | 0.65 |
| 0.2 | 0.65 |
| 0.3 | 0.63 |
| 0.4 | 0.66 |
| 0.5 | 0.63 |

Table 10 shows that there is no fixed pattern for the change of AP with different values of the overlapping area ratio. Overall, as the value of the overlapping area ratio increases, AP also increases. However, after reaching a certain threshold, AP starts to decrease. This is because when the value of the overlapping area ratio is too small, some aphids may not be detected as they are split into different parts with low classification confidence. Conversely, when the value of the overlapping area ratio is too large, aphids in the overlapping areas between different blocks will be detected multiple times, resulting in over-detection.

To sum up, the optimal value of the size of the slicing window or the overlapping area ratio is determined by the distribution density of aphids in the dataset.

*3.6 Setting of the threshold used to switch between different aphid counting networks*

Our proposed automatic aphid counting network can automatically select different counting networks in term of the distribution density of aphids in the image. When the distribution density of aphids is lower than a pre-defined threshold T, the improved Yolov5 is employed to count aphids. On the other hand, when the distribution density of aphids exceeds the threshold T, it switches to CSRNet for counting aphids. To investigate the optimal threshold T, we conducted corresponding test experiments. Specifically, we varied T from 0 to 200 with an interval of 5 and carried out aphid counting test on the test set of the standard aphid dataset Aphid_Dataset_1. The MAE and RMSE change curves with different thresholds are shown in Fig. 14 and Fig. 15.

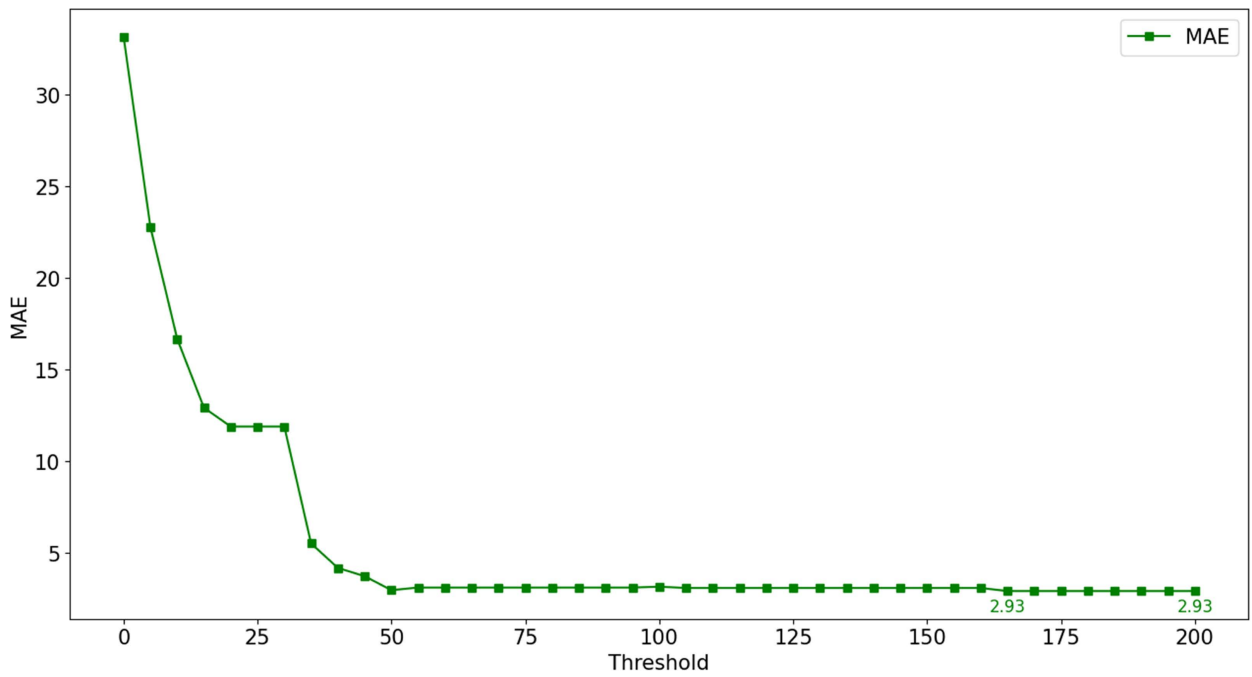

**Fig. 14.** The MAE change curve with different thresholds.

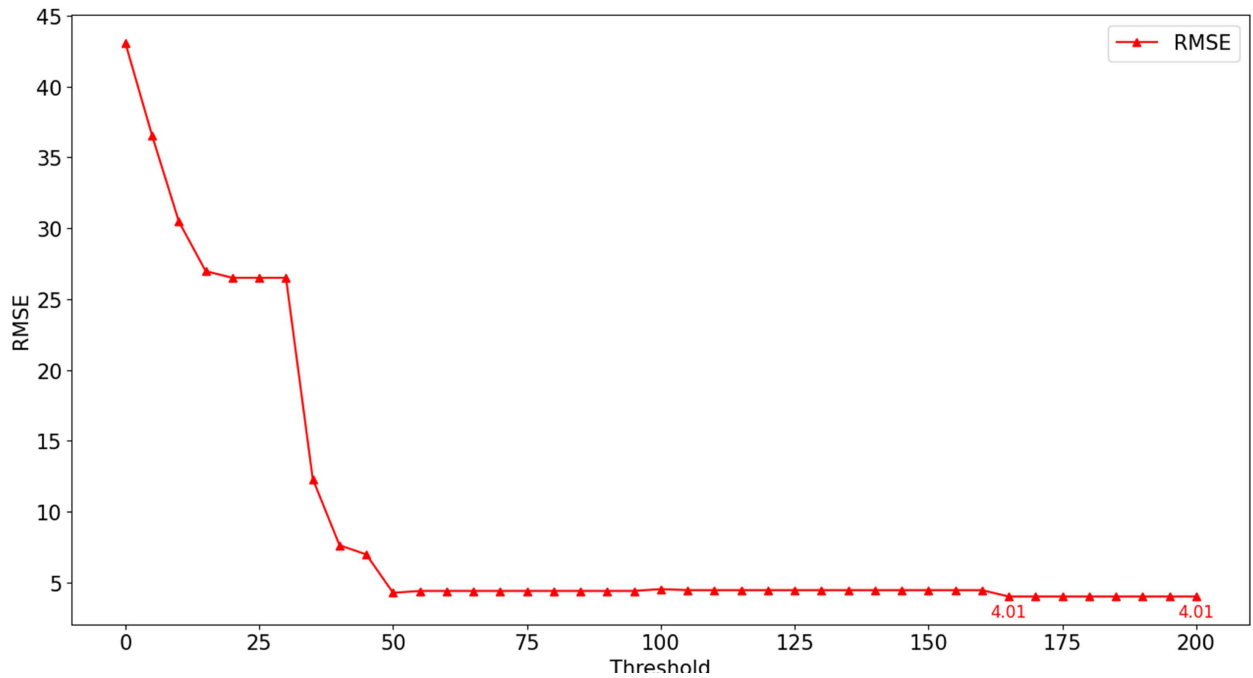

**Fig. 15.** The RMSE change curve with different thresholds.

From Fig. 14 and Fig. 15, it can be seen that the change trends of MAE and RMSE under different thresholds are basically consistent. Initially, both MAE and RMSE decrease rapidly with increasing T. However, after T reaches 50, the changes in MAE and RMSE gradually stabilize. When T is within the range of [165, 200], the values of both MAE and RMSE reach their minimums, with 2.93 and 4.01, respectively.

## 4. Discussion

We evaluated accuracy across low-density and high-density scenarios. The overestimated test accuracy phenomenon, common in many works, arises from cross-duplication in training and test sets and overly idealized experimental setups. Our work avoids this through careful dataset segregation and data augmentation limited to the training set. In comparing automatic aphid counting methodologies, variations in datasets and evaluation metrics make direct comparisons challenging, but some trends can still be discerned. Basically, there are two testing environments which are direct counting on crop leaves and counting in traps. Direct counting provides a realistic representation of aphid population dynamics, albeit at the expense of accuracy due to challenges like leaf occlusion. Traps offer greater accuracy but only capture winged adult aphids, hence providing a relative measure of the population (Liu et al., 2016).

The choice of data acquisition device can significantly affect counting accuracy. High-resolution devices enable detailed aphid morphology representation, but their high cost, need for specialized customization, and requirement of indoor experimental environments complicate their use (Júnior et al., 2022; Kalfas et al., 2023). Conversely, smartphones offer a convenient alternative but compromise on resolution, affecting aphid feature extraction (Li et al., 2019).

Many studies noted a decrease in counting accuracy with increasing aphid distribution density without addressing the issue. Our work differs in developing a hybrid network to handle varied density scenarios, combining detection networks for low-density and density map estimation networks for high-density populations. Despite our innovative approach, high-density aphid counting remains a significant challenge. Notably, unlike other studies, we used multiple images for each density level, ensuring that our conclusions are more representative and reliable.

Reflecting on our work in this paper, there is still room for improvement. Firstly, our proposed automatic aphid counting approach requires manual data collection in extremely harsh agricultural environments, which still involves a certain level of labor-intensive efforts. To address this, an automated aphid data collection station with the functions of controllable filling and emptying water is highly needed to automate the whole pipeline of aphid counting. Labelling tiny aphids is a tedious process, therefore semi-automatic labeling or weakly supervised learning could be exploited to avoid labelling large datasets for fully supervised deep learning models. Secondly, through the comparison of results from Table 5 and Table 7, as well as Table 6 and Table 8, we have identified that augmenting the dataset can, to some extent, improve the testing accuracy of the model. Therefore, it is essential for future work to further expand the aphid dataset. Thirdly, the accuracy of our proposed hybrid network for automatic aphid counting requires further improvement in future research. In the detection network branch, although the improved Yolov5 demonstrates a 5% increase in AP compared with the original Yolov5, it achieves only 65% AP on the test set of the standard aphid dataset (Table 4). The most significant challenge in improving the accuracy of aphid counting lies in the tiny size of aphids in the images, which makes it difficult to extract detailed features. Since the low requirement for real-time performance in automatic aphid counting, a larger convolution kernel instead of the commonly used small kernels ($3 \times 3$, $1 \times 1$) might be helpful to fine tune feature extraction. We argue that a higher-resolution image with clear aphid details is directly way to improve the results as it aims to enhance the model's ability to capture fine-grained features of aphids and improve the overall accuracy of aphid counting. We found that the size of the slicing window, the overlapping area ratio in the detection network branch, and the threshold T used to switch between different aphid counting networks significantly affect the final aphid counting accuracy (Table 9, Table 10, Fig. 14 and Fig. 15). And the optimal values for these parameters depend on the distribution density of aphids in the dataset. Therefore, how to adaptively select the optimal values of the above parameters for different aphid datasets is a significant step towards robust and accurate aphid counting.

## 5. Conclusion

The novelty in this paper lies primarily in the proposal of a hybrid automatic aphid counting network architecture that integrates the detection network (an improved Yolov5) and the density map estimation network (CSRNet). It automatically switches between two network branches for accurate counting of aphids according to the distribution density of aphids in the images, resulting in the best counting performance across various aphid distribution conditions. We have verified that our proposed automatic aphid counting network is superior to other methods through comparison experiments. It achieved the lowest MAE and RMSE values across both the standard and high-density aphid datasets. Additionally, the improved Yolov5 demonstrated better performance than the original Yolov5. We also conducted ablation experiments to further investigate the effectiveness of each improvement component in the improved Yolov5 and the impact of dataset size on model testing accuracy. The results showed the best results were achieved when the components were combined rather than used independently, and expanding the dataset can improve the testing accuracy of the model to some extent. Finally, we further obtained the optimal values for three important parameters through experiments, including the size of the slicing window, the overlapping area ratio in the detection network branch, and the threshold T used to switch between different aphid counting networks. At the same time, we concluded that these parameters are influenced by the distribution density of aphids in the dataset, which ultimately affects the accuracy of aphid counting. This work provides an advanced solution that can be widely applied to the automatic counting and monitoring of all small pests in agricultural environments. Further research is required to expand the aphid dataset, automate data collection, improve the network's counting accuracy and adaptability across different datasets.

**Funding**

This work was supported by the Engineering and Physical Sciences Research Council

[EP/S023917/1], the AgriFoRwArdS CDT, the BBRO, and Lincoln Agri-Robotics. This work was also partly supported by the International cooperation foundation of tobacco research institute of Chinese Academy of Agricultural Sciences No. IFT202303, Major special projects for green pest control (110202001035(LS-04)) and Major special projects for big data (110202201051(SJ-01)).

**CRediT authorship contribution statement**

**Xumin Gao:** Investigation, Data curation, Methodology, Software, Validation, Visualization, Writing - original draft, Writing - review & editing. **Wenxin Xue:** Writing - review & editing. **Mark Stevens:** Data curation, Writing - review & editing, Funding acquisition. **Callum Lennox:** Writing - review & editing. **Junfeng Gao:** Conceptualization, Supervision, Formal analysis, Project administration, Writing - review & editing, Funding acquisition.

**Declaration of Competing Interest**

The authors declare that they have no known competing financial interests or personal relationships that could have appeared to influence the work reported in this paper.

**Data availability**

The dataset and codes can be accessed with this link https://github.com/JunfengGaolab/Counting-Aphids.